\documentclass[letterpaper, 10 pt, conference]{ieeeconf}  

\IEEEoverridecommandlockouts                              

\usepackage{graphics} 
\usepackage{epsfig} 
\usepackage{mathptmx} 
\usepackage{times} 
\usepackage{amsmath} 
\usepackage{amssymb}  
\usepackage{booktabs}
\usepackage{makecell}
\usepackage{xcolor} 
\usepackage{multirow}
\usepackage{amssymb}
\usepackage{ragged2e}
\usepackage{pifont}

\usepackage{tabularx}

\newcommand\our{RoboPCA}
\newcommand\ourdata{Human2Afford}

\title{\LARGE \bf
\our{}: Pose-centered Affordance Learning from Human Demonstrations for Robot Manipulation
}

\author{Zhanqi Xiao, Ruiping Wang and Xilin Chen
\thanks{*This work is partially supported by Beijing Municipal Natural Science Foundation Nos. L257009, L242025, and Natural Science Foundation of China under contracts Nos. 62495082, 62461160331.}
\thanks{The authors are with the Key Laboratory of AI Safety of CAS, Institute of Computing Technology, Chinese Academy of Sciences (CAS), Beijing, 100190, China, and also with the University of Chinese Academy of Sciences, Beijing 100049, China. \tt\footnotesize\justifying{\{zhanqi.xiao\}@vipl.ict.ac.cn, \{wangruiping,xlchen\}@ict.ac.cn}}
\thanks{Corresponding author: Ruiping Wang.}
}

\begin{document}

\maketitle
\thispagestyle{empty}
\pagestyle{empty}

\begin{abstract} 
Understanding spatial affordances---comprising the contact regions of object interaction and the corresponding contact poses---is essential for robots to effectively manipulate objects and accomplish diverse tasks. However, existing spatial affordance prediction methods mainly focus on locating the contact regions while delegating the pose to independent pose estimation approaches, which can lead to task failures due to inconsistencies between predicted contact regions and candidate poses. In this work, we propose \our{}, a pose-centered affordance prediction framework that jointly predicts task-appropriate contact regions and poses conditioned on instructions. To enable scalable data collection for pose-centered affordance learning, we devise \ourdata{}, a data curation pipeline that automatically recovers scene-level 3D information and infers pose-centered affordance annotations from human demonstrations. With \ourdata{}, scene depth and the interaction object's mask are extracted to provide 3D context and object localization, while pose-centered affordance annotations are obtained by tracking object points within the contact region and analyzing hand–object interaction patterns to establish a mapping from the 3D hand mesh to the robot end-effector orientation. By integrating geometry–appearance cues through an RGB-D encoder and incorporating mask-enhanced features to emphasize task-relevant object regions into the diffusion-based framework, \our{} outperforms baseline methods on image datasets, simulation, and real robots, and exhibits strong generalization across tasks and categories.

\end{abstract}

\section{INTRODUCTION}



\begin{figure*}[!htbp]
    \centering
    \includegraphics[width=1.0\textwidth]{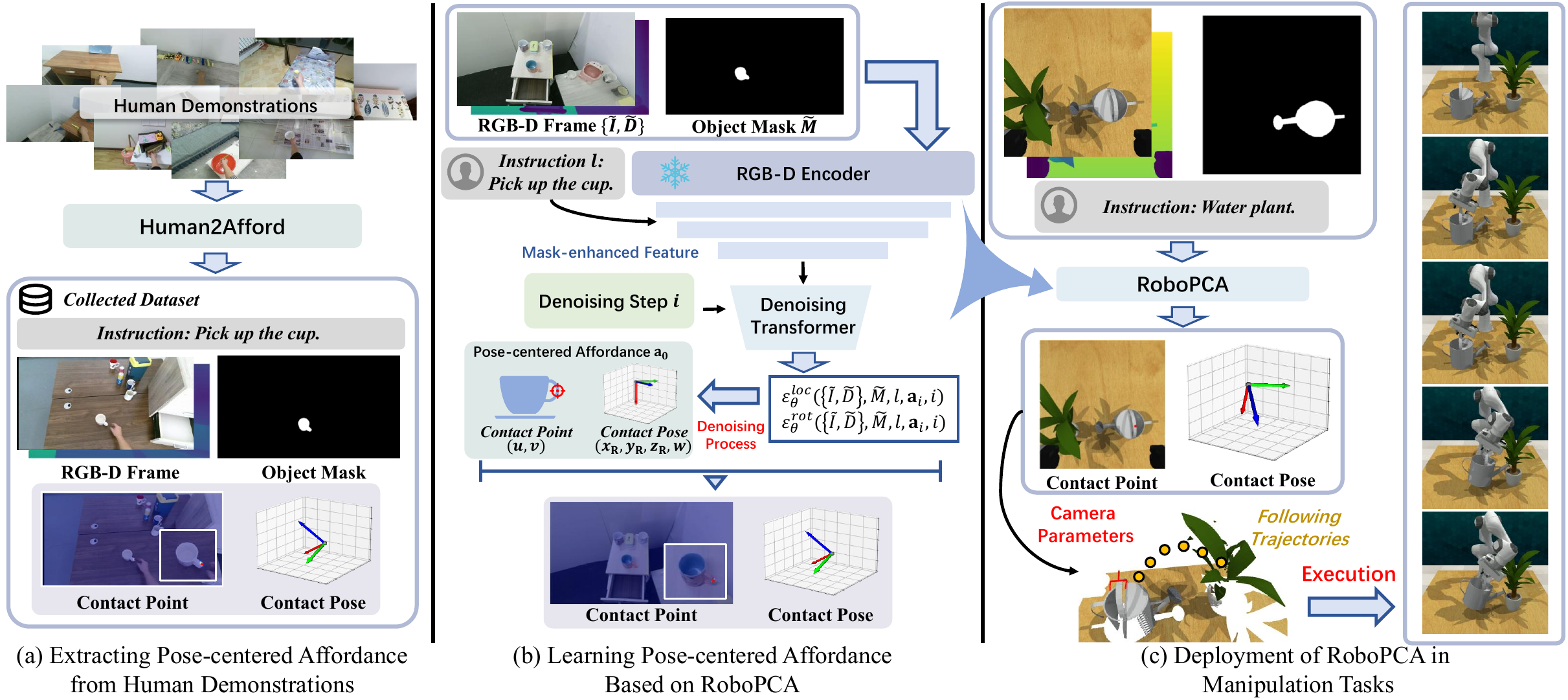}
    \vspace{-1em}
    \caption{Overview of our pipeline. (a) Pose-centered affordance annotations and complementary scene information are extracted from human demonstrations with \ourdata{} for pose-centered affordance learning. (b) \our{} builds upon a diffusion framework to predict pose-centered affordances, an RGB-D encoder is used to effectively capture both geometry and appearance cues, and mask-enhanced features are incorporated to emphasize task-relevant object regions. (c) The predicted pose-centered affordances are transformed into 6-DoF poses using camera parameters, guiding the robot to complete the task.}
    \vspace{-1em}
    \label{fig:overview}
\end{figure*}

Humans excel at manipulating various objects in unstructured environments, including locating task-related interaction regions and selecting appropriate contact poses. This ability is largely ascribed to the profound understanding of spatial affordances~\cite{gibson2014ecological, gibson1966senses}.  As robots and humans operate in the same workspace, a comprehensive understanding of spatial affordances is crucial for enhancing robotic manipulation capabilities across diverse object categories and tasks.

Affordance had already been extensively studied before its application to robotic manipulation, which is often represented as masks or heatmaps~\cite{qian2024affordancellm, tang2024uad, ma2024glover}. While such representation can indicate potential interaction regions on objects, it lacks the precise spatial localization required for accurate robot manipulation. To address this, recent works propose representing affordances in terms of contact points and post-contact trajectories~\cite{bahl2023affordances}. In this representation, the interaction region is denoted by a 2D pixel to specify where the robot should make contact with the object, while the post-contact trajectories are represented by a 2D vector on the image to indicate the direction and movement the robot should execute after making contact. Building on this representation, subsequent works have made many improvements, such as introducing object retrieval or leveraging the reasoning capability of foundation models for generalization across categories~\cite{ju2024robo, kuangram, liu2024moka, yuan2025robopoint, ma2025glover++}. However, contact points and post-contact trajectories alone do not specify the manipulator pose for task execution. Although the final pose could be obtained by filtering the grasp candidates from independent pose estimation approaches~\cite{fang2023anygrasp, tang2025foundationgrasp} based on the predicted contact point, the inconsistency between predicted contact points and grasp candidates may lead to suboptimal or even failed executions. We argue \textbf{pose-centered affordances}, represented jointly by the \textbf{contact points} and the corresponding \textbf{contact poses}, provide a more coherent and expressive formulation for robotic manipulation. By unifying contact localization and pose estimation, this formulation reduces the inconsistency between predicted contact points and grasp candidates and provides a principled basis for generating reliable manipulation strategies.

Similar to other tasks in robot learning, learning pose-centered affordances that support the manipulation of arbitrary objects requires large-scale data. While recent works, such as DROID~\cite{khazatsky2024droid}, have gathered large-scale robotic demonstrations via human teleoperation, this approach remains difficult to scale to new environments or task demands. In contrast, human demonstrations provide a promising data source, not only due to the scale and diversity of scenes and tasks, but also because they naturally capture dynamics relevant to object interactions. Nevertheless, the absence of 3D information and low-level action labels limits their utility for pose-centered affordance learning, particularly for extracting contact poses.

In this work, we aim to enable pose-centered affordance learning from a large amount of unlabeled human demonstrations, tackling two key questions: (1) How can pose-centered affordances be extracted from unlabeled human demonstrations? (2) How to effectively learn pose-centered affordances from collected datasets? To answer the first question, we devise \textbf{\ourdata{}}, a data curation pipeline that automatically recovers scene-level 3D information and extracts pose-centered affordances from human demonstrations. With each identified pre-contact frame and contact frame pair, \ourdata{} first recovers the depth information and extracts the interaction object's mask in the pre-contact frame to provide 3D context and object localization to facilitate pose-centered affordance learning. To obtain pose-centered affordance annotations, we first recover the contact pose by analyzing the hand–object interaction patterns to establish a mapping between the estimated 3D hand mesh in the contact frame and the robot end-effector’s orientation, and then derive the contact point by tracking object pixels within the contact region in the contact frame back to the pre-contact frame. We collected 10K human–object interaction images with pose-centered affordance annotations and complementary scene information using \ourdata{}. Based on the collected dataset, we propose \textbf{\our{}} for the second question, which is a \textbf{P}ose-\textbf{C}entered \textbf{A}ffordance prediction framework that jointly predicts task-appropriate contact point and corresponding contact pose given instructions. \our{} builds upon a diffusion framework to predict pose-centered affordances. To effectively capture both geometry and appearance cues, it leverages a state-of-the-art RGB-D encoder~\cite{yin2025dformerv2} that integrates color and depth information. Additionally, mask-enhanced features are incorporated to emphasize task-relevant object regions, improving the model’s ability to localize interaction points and accurately infer corresponding contact poses. The overview of our pipeline is shown in Fig.~\ref{fig:overview}.

We evaluate \our{} through extensive experiments on image datasets, simulation, and real-world scenarios. It outperforms baseline methods in contact point prediction precision and manipulation task success rates, achieving improvements of 18.6\% on AGD20K~\cite{luo2022learning} evaluated on object categories feasible for robotic manipulation, 38.5\% on RLBench~\cite{james2020rlbench}, and 24.9\% in real-world experiments, which demonstrates that \our{} not only achieves higher precision in predicting contact points but also provides more reliable and effective guidance for manipulation tasks. We also demonstrate its compatibility with robotic data.

\section{RELATED WORK}

\begin{figure*}[t]
    \centering
    \includegraphics[width=1\textwidth]{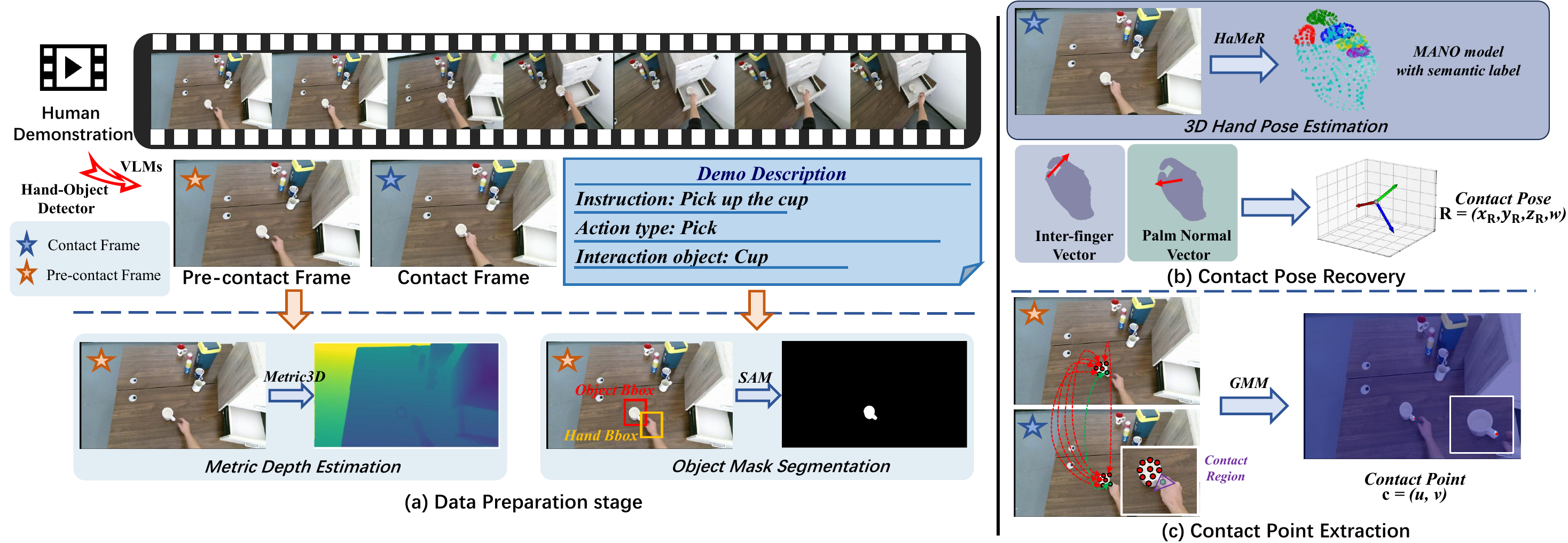}
    \vspace{-2em}
    \caption{Overview of \ourdata{}. (a) Given a human demonstration, we identify the demo description and extract key frames using a hand–object detector and VLMs. Depth and the interaction object mask are then obtained via metric depth estimation and segmentation. (b) Using the 3D hand mesh from a hand pose estimator, we extract the contact pose based on the inter-finger vector and palm normal. (c) Object points are tracked from the pre-contact to the contact frame, and points within the inter-finger contact region are modeled with GMM to extract the contact point.}
    \vspace{-1em}
    \label{fig:data_curation}
\end{figure*}

\subsection{Visual Affordance Learning}

Affordance focuses on determining where and how to interact with diverse objects based on the visual inputs and instructions. One line of work learns affordances from annotated datasets~\cite{delitzas2024scenefun3d, do2018affordancenet}, which is often prohibitively expensive. To alleviate the labeling cost, some studies learn affordances by exploring effective interaction in simulated environments~\cite{geng2023partmanip, mo2021where2act}; however, acquiring diverse virtual assets also incurs substantial cost. Others leverage large-scale models to automatically annotate contact regions~\cite{tang2024uad, ma2024glover}, but these annotations are typically performed on single-frame images or static assets and thus fail to capture the dynamic information about how to interact with objects. 

In contrast, human demonstrations have gained increasing attention as a more general source for affordance learning, as they naturally encode interaction dynamics and are abundantly available online, offering rich scene and task diversity. Building upon this formulation proposed by~\cite{bahl2023affordances}, which represents affordance as contact points and post-contact trajectories by a 2D pixel and a 2D vector on the image, object retrieval has been employed to enable cross-category transfer~\cite{ju2024robo, kuangram}, and knowledge from large-scale models has been leveraged to enhance the robustness of contact point prediction~\cite{liu2024moka, yuan2025robopoint, ma2025glover++}. With respect to post-contact trajectories, recent studies have extended the 2D vector representation into point-based trajectories to capture curved or non-linear motions that cannot be expressed in a single vector, thereby offering a more faithful description of complex interaction dynamics~\cite{chen2025vidbot, xu2025a0}. However, these approaches require an independent pose estimation module to generate the final contact pose by filtering, which could lead to suboptimal or failed execution due to inconsistencies between predicted contact points and grasp candidates. In contrast, our model jointly predicts contact points and corresponding poses to provide a more consistent representation for robot manipulation.

\subsection{Robot Learning from Humans}

Human demonstrations offer a natural and abundant source of supervision for robotic learning, as they inherently capture the dynamics of interaction. With the availability of large-scale video datasets~\cite{liu2022hoi4d, damen2018scaling, hoque2025egodex}, recent studies have explored utilizing human videos to facilitate robot learning. One line of approaches learns visual representations from human videos and leverages the pre-trained visual encoders to facilitate policy network training~\cite{nair2022r3m, xu2024flow, bharadhwaj2024towards, srirama2024hrp}. Another line of research focuses on learning reward functions from human videos~\cite{chen2021learning, chang2023look,  shaw2023videodex}. Additionally, some works leverage the motion attributes extracted from human videos, such as wrist trajectories or 3D hand poses, to guide policy learning or to serve as intermediate supervision for manipulation tasks~\cite{wang2023mimicplay, bharadhwaj2024gen2act, haldar2025point}. Building upon these directions, affordance provides an explicit and interpretable representation that can naturally transfer from human to robot embodiment. In this work, our model provides a principled basis for generating reliable, context-aware manipulation strategies through learning pose-centered affordances that jointly predict contact points and contact poses directly from human demonstrations.

\subsection{Diffusion Models in Robotics}

Diffusion models have emerged as a powerful paradigm for modeling complex data distributions through iterative denoising, and have also demonstrated strong potential in robotic learning as effective frameworks for policy learning~\cite{janner2022planning, chi2023diffusion, ze20243d, ke20243d, liang2024skilldiffuser}. Diffusion Policy~\cite{chi2023diffusion} provides a general framework for generating robot action trajectories via a conditional denoising diffusion process. Building on this, subsequent works have explored enhancing trajectory generation by incorporating reward signals to guide the denoising process~\cite{janner2022planning}, while other approaches adopt a more factorized policy learning framework that unifies action keypose prediction and trajectory diffusion generation for learning robot manipulation from demonstrations~\cite{ke20243d, liang2024skilldiffuser}. However, these approaches often suffer from limited generalization ability, making it difficult to transfer across diverse objects and environments. Instead of predicting full trajectories, our model uses diffusion models for pose-centered affordance learning. Conditioned on 3D scenes' information and instructions, it jointly generates contact points and poses that generalize across diverse tasks and object types.

\section{METHOD}

\subsection{Problem Formulation}

We aim to learn a pose-centered affordance prediction model $\mathbf{a} = \pi(\{\tilde{I}, \tilde{D}\}, l, \tilde{M})$ from human demonstrations, where $\{\tilde{I}, \tilde{D}\}$ denotes an RGB-D frame (with image $\tilde{I}$ and depth $\tilde{D}$), $l$ represents the language instruction and $\tilde{M}$ is the target object's mask relevant to the instruction. The depth frame could be obtained either from a depth sensor or a metric-depth estimation foundation model~\cite{hu2024metric3d}, and the target object mask can be easily acquired through integrating an open-vocabulary object detection method~\cite{liu2024grounding} and segmentation foundation models~\cite{ravi2024sam}. We formulate the pose-centered affordance representation $\mathbf{a}$ in terms of contact points $\mathbf{c}$, represented as 2D points in pixel space, and end-effector orientations $\textbf{R}$, represented as quaternions in 3D space relative to the camera coordinate frame. Specifically, $\mathbf{a} = \{\mathbf{c}, \mathbf{R}\}$, where $\mathbf{c} = (u, v)$ and $\mathbf{R} = (w, x_\mathbf{R}, y_\mathbf{R}, z_\mathbf{R})$, with $(w, x_\mathbf{R}, y_\mathbf{R}, z_\mathbf{R})$ denoting the quaternion components. Using the camera intrinsic parameters $\mathbf{K}=(f_x, f_y, c_x, c_y)$ and the depth value $\tilde{D}_{u,v}$ at pixel $(u,v)$, the 3D contact position $\mathbf{p} = (x, y, z)$ can be computed as:
\begin{equation}
\begin{gathered}
    z = \tilde{D}_{u,v}, \quad
    x = (u - c_x) \cdot \frac{z}{f_x}, \quad
    y = (v - c_y) \cdot \frac{z}{f_y}.
\end{gathered}
\end{equation}

The final 6-DoF pose, which the robot uses to interact with the target object under the given instruction, is then represented as $\mathbf{\tau} = (\mathbf{p}, \mathbf{R}) = (x, y, z, w, x_\mathbf{R}, y_\mathbf{R}, z_\mathbf{R}).$ To enhance clarity in subsequent discussions, we will use $\mathbf{R}$ to directly represent the contact pose.

\subsection{Pose-centered Affordance Extraction from Human Videos}

Human videos are often captured with moving monocular cameras, where the camera pose and depth per frame are unknown. The absence of low-level action labels and 3D scene information impedes robots from acquiring manipulation skills directly from human demonstrations. To enable learning pose-centered affordances from human videos, we devise \textbf{\ourdata{}} that automatically recovers scene-level 3D information and extracts pose-centered affordances from human demonstrations, as shown in Fig.~\ref{fig:data_curation}. Here, we introduce the key components in our data curation pipeline.

\textbf{Data Preparation.}
\label{sec:data_pre}
Given a video clip $V$ consisting of $T$ frames i.e. $V=\{I_1, \dots, I_T\}$, depicting human–object interactions (e.g., a person picking up a cup), we first prompt a vision-language model (VLM), specifically Gemini-2.0-Flash~\cite{team2024gemini}, to identify the action depicted and the category of the interacting object. To extract pose-centric affordances from human videos, we first identify the contact frame where the interaction occurs and the pre-contact frame where the target object remains unoccluded. We utilize a widely-adopted hand-object interaction detector~\cite{shan2020understanding} to identify the interaction state in each frame and obtain the corresponding hand bounding boxes. Since significant changes in camera pose between the pre-contact frame and the contact frame can hinder 3D pose recovery due to the lack of camera parameter labels, we further leverage Gemini to examine frames around the critical interval of interaction state changes, filtering out false positives and selecting contact and pre-contact frames $\{I_c, I_p\}$ that are temporally close to satisfy the viewpoint consistency requirement. We then employ a metric-depth foundation model~\cite{hu2024metric3d} to recover the depth $\hat{D}$ of $I_p$, and use GroundingDINO~\cite{liu2024grounding} and SAM2~\cite{ravi2024sam}, guided by hand bounding box priors, to obtain the interacting object's mask in $I_c$, which is then projected back onto $I_p$, denoted as $\hat{M}$.


\textbf{Contact Pose Recovery.}
Under the assumption that there is no significant change in camera pose between $I_p$ and the $I_c$, we first leverage a 3D hand pose estimator~\cite{pavlakos2024reconstructing} to estimate the hand pose in $I_c$. With the hand mesh given by~\cite{pavlakos2024reconstructing}, we propose a heuristic method that establishes a mapping from the recovered human hand pose to the robot end-effector orientation.
Since human-object interactions predominantly involve the thumb, index, and middle fingers, we identify the finger pair that primarily applies force to the object by jointly analyzing the inter-finger mesh distances and their spatial relationships with the object mask $\hat{M}$. Based on the inter-finger vector of the selected finger pair $\hat{\mathbf{v}}_{fp}$ and the averaged normal vector of the palm region $\hat{\mathbf{n}}_{palm}$, we recover the robot contact pose $\mathbf{R}_c$ from the human hand pose, here $\mathbf{R}_c$ is aligned with the orientation $\mathbf{R}$ in the $I_p$, denoted as:
\begin{equation}
    \mathbf{R} = \mathbf{R}_c = \mathcal{F}(\hat{\mathbf{v}}_{fp}, \hat{\mathbf{n}}_{palm}).
\end{equation}

\textbf{Contact Point Extraction.}
To extract the contact point $c$, we employ SpaTracker~\cite{xiao2024spatialtracker}, an off-the-shelf dynamic point tracker, to automatically track the interaction object from $I_p$ to $I_c$, providing temporally consistent object localization even under occlusion. We then analyze the overlap between the object points and the hand mesh to identify points within the area formed by the thumb, index, and middle fingers in $I_c$. These points are back-projected to $I_p$, denotes as $\{\mathbf{c}_i\}^N$. Treating $p(\mathbf{c})$ as the distribution over $\{\mathbf{c}_i\}^N$, we fit a Gaussian mixture model (GMM) with parameters $(\mathbf{\mu}_k, \mathbf{\sigma}_k)$ by maximizing the likelihood over all $\mathbf{c}_i$:
\begin{equation}
    p(\mathbf{c})=\mathop{argmax}\limits_{\mathbf{\mu}_1, ...,\mathbf{\mu}_K, \mathbf{\sigma}_1,....\mathbf{\sigma}_K} \sum\limits_{i=1}^N \sum\limits_{k=1}^K \, \mathcal{N}(\mathbf{c}_i \mid \mathbf{\mu}_k, \mathbf{\sigma}_k).
\end{equation}
The final contact point in $I_p$ is represented as the average of the Gaussian means:
\begin{equation}
    \mathbf{c} = \frac{1}{K} \sum\limits_{k=1}^K \ \mathbf{\mu}_k
\end{equation}

\subsection{Pose-centered Affordance Learning}
\our{} is trained as a conditional diffusion probabilistic model to infer pose-centered affordance $\mathbf{a}=\{\mathbf{c}, \mathbf{R}\}$ given the RGB-D frame of the scene $\{\tilde{I}, \tilde{D}\}$, target object mask $\hat{M}$, and a language instruction $l$ through iterative denoising process. We represent rotations $\mathbf{R}$ using the 6D rotation representation of~\cite{zhou2019continuity} to avoid the discontinuities of the quaternion representation. A variance schedule $\{\beta_i \in (0, 1)\}_{i=1}^N$ is associated with the diffusion process, which defines how much noise is added at each diffusion step. Given a sample $\mathbf{a}_0$, the noise version of $\mathbf{a}_0$ at step \textit{i} can be written as $\mathbf{a}_i = \sqrt{\bar{\alpha}_i} \mathbf{a}_0 + \sqrt{1-\bar{\alpha}_i}\mathbf{\epsilon}$, where $\mathbf{\epsilon} \sim \mathcal{N}(0, 1)$ is a sample noise from a Gaussian distribution with the same dimension as $\mathbf{a}_0$, $\bar{\alpha_i} = \prod\limits_{j=1}^i \alpha_i$ and $\alpha_i = 1-\beta_i$.

\our{} models a learned gradient of the denoising process with a denoising transformer $\hat{\mathbf{\epsilon}} = \mathbf{\epsilon}_\theta(\mathbf{a}_i; i, \{\tilde{I}, \tilde{D}\}, l, \tilde{M})$ that takes the noisy pose-centered affordance $\mathbf{a}_i$, diffusion step $i$, and conditioning information from the RGB-D frame of current scene $\{\tilde{I}, \tilde{D}\}$, object mask $\tilde{M}$ and language instruction $l$ as input, to predict the noise component $\hat{\mathbf{\epsilon}}$.

At each diffusion step \textit{i}, we convert the visual observation of the scene $\{\tilde{I}, \tilde{D}\}$ and noised pose-centered affordance estimate $\mathbf{a}_i$ to a set of tokens, where each token is represented as a latent embedding and a position in pixel coordinates. Since human demonstrations lack camera pose annotations, we incorporate geometric information, which is crucial for affordance understanding, by leveraging a state-of-the-art RGB-D encoder~\cite{yin2025dformerv2} to encode each RGB-D frame. Moreover, with the object mask $\tilde{M}$, we deploy the same encoder to the masked RGB-D frame to enhance the model's perception of task-relevant object regions. The features from both the full and masked frames are then concatenated to obtain the mask-enhanced features of the scene. The noisy estimate $\mathbf{a}_i$ of the clean pose-centered affordance $\mathbf{a}_0$ is transferred to a latent embedding vector with an MLP, and the language task instruction is mapped to language tokens using a pre-trained CLIP language encoder. We fuse tokens of scene and $\mathbf{a}_i$ by applying relative self-attentions among all tokens, and additionally fuse language tokens using cross-attentions to get the final conditional tokens. We use the rotary positional embeddings~\cite{kitaev2020reformer} to encode relative positional information in attention layers. The conditional tokens is then fed into MLPs to predict the noise $\mathbf{\epsilon}_\theta^{loc}(\{\tilde{I}, \tilde{D}\}, \tilde{M}, l, \mathbf{a}_i, i)$ and $\mathbf{\epsilon}_\theta^{rot}(\{\tilde{I}, \tilde{D}\}, \tilde{M}, l, \mathbf{a}_i, i)$ added to $\mathbf{a}_0$'s contact point $\mathbf{c}_0$ and contact pose $\mathbf{R}_0$.

During training stage, we randomly sample a diffusion step $i$ and add noise $\mathbf{\epsilon} = (\mathbf{\epsilon}^{loc}, \mathbf{\epsilon}^{rot})$ to ground truth pose-centered affordance $\mathbf{a}_0 = (\mathbf{c}_0, \mathbf{R}_0)$. We use L1 loss for the prediction of $\mathbf{\epsilon}$. The objective loss of the denoising transformer $\mathcal{L}_{\theta}$ can be expressed as \eqref{eq:denoise_loss}, where $\omega_1$ and $\omega_2$ are hyperparameters controlling the relative weights of $\mathcal{L}_{\theta}^{loc}$ and $\mathcal{L}_{\theta}^{rot}$.
\begin{equation} \label{eq:denoise_loss}
\begin{split}
    \mathcal{L}_{\theta}^{loc} &= \| \epsilon_{\theta}^{loc}(\{\tilde{I}, \tilde{D}\}, \tilde{M}, l, \mathbf{a}_i, i) - \epsilon^{loc} \|, \\
    \mathcal{L}_{\theta}^{rot} &= \| \epsilon_{\theta}^{rot}(\{\tilde{I}, \tilde{D}\}, \tilde{M}, l, \mathbf{a}_i, i) - \epsilon^{rot} \|, \\
    \mathcal{L}_{\theta} &= \omega_1 \cdot \mathcal{L}_{\theta}^{loc} + \omega_2 \cdot \mathcal{L}_{\theta}^{rot}.
\end{split}
\end{equation}

During inference, a sample $\mathbf{a}_N \sim \mathcal{N}(0,1)$ is first drawn. The predicted contact point $\hat{\mathbf{c}}$ and contact pose $\hat{\mathbf{R}}$ are obtained through progressively denoising the sample $N$ times with $\epsilon_{\theta}$, following \eqref{eq:denoise_eq}, where $\mathbf{z} \sim \mathcal{N}(0,1)$ is a random variable of appropriate dimension. Scaled-linear and square cosine scheduler are used for $\mathbf{c}$ and $\mathbf{R}$ seperately to achieves better performance~\cite{ke20243d}.

\begin{equation} \label{eq:denoise_eq}
\resizebox{0.91\columnwidth}{!}{$
    \mathbf{a}_{i-1} = \frac{1}{\sqrt{\alpha_i}}\left(\mathbf{a}_i - \frac{\beta_i}{\sqrt{1-\bar\alpha_i}} \epsilon_{\theta}(\{\tilde{I}, \tilde{D}\}, \tilde{M}, l, \mathbf{a}_i, i)\right) + \frac{1-\bar\alpha_{i+1}}{1-\bar\alpha_i}\beta_i \mathbf{z}
$}
\end{equation}

\section{EXPERIMENTS}
\begin{figure}[t]
    \centering
    \includegraphics[width=0.5\textwidth]{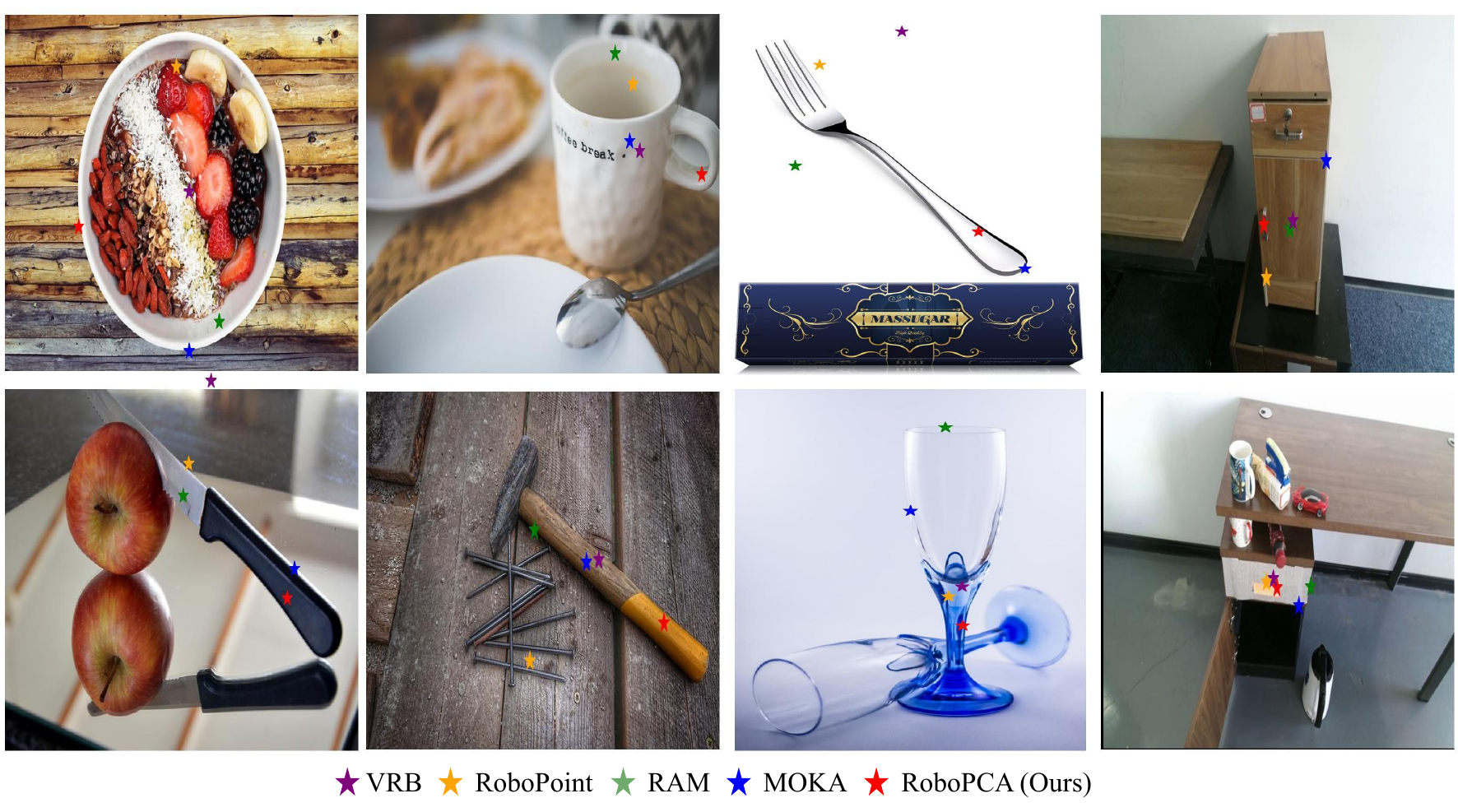}
    \vspace{-2em}
    \caption{Qualitative results on AGD20K. The label {\large\textcolor{purple}{$\star$} \textcolor{orange}{$\star$} \textcolor{green}{$\star$} \textcolor{blue}{$\star$} \textcolor{red}{$\star$}} indicate the predicted contact points of different methods.}
    \vspace{-1em}
    \label{fig:AGD20K}
\end{figure}

\begin{figure*}[t]
    \centering
    \includegraphics[width=1\textwidth]{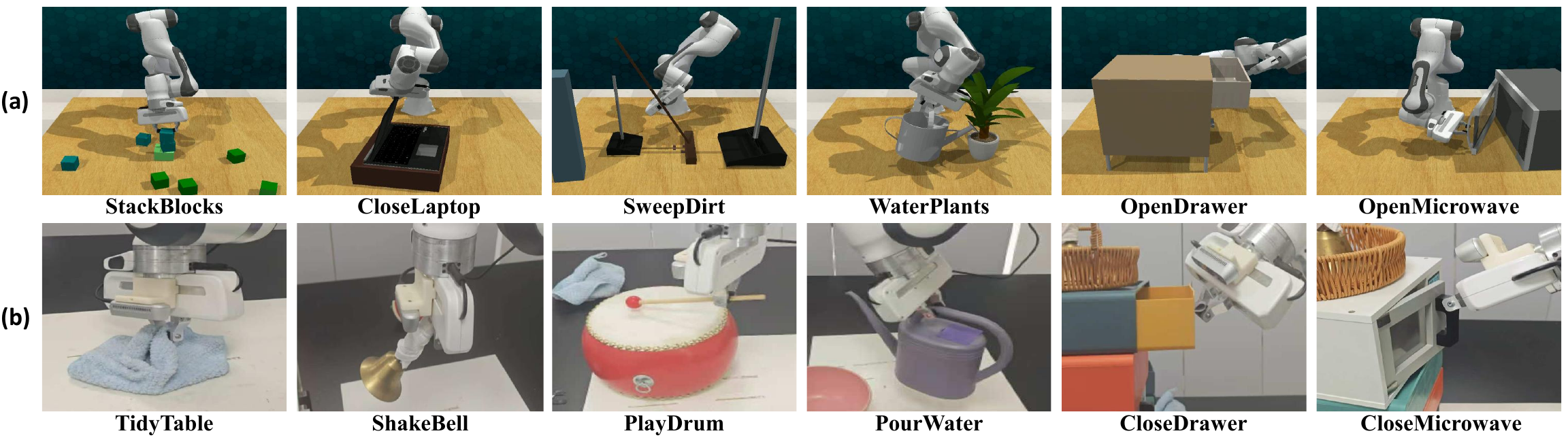}
    \vspace{-2em}
    \caption{Examples of task settings in simulation and the real world. We evaluate our method on 10 tasks in simulation (a; only 6 shown), and 9 tasks in the real world (b; only 6 shown) across various object categories to validate its effectiveness.}
    \label{fig:task_setting}
    \vspace{-0.5em}
\end{figure*}

To verify the effectiveness of the \our{}, we conduct extensive experiments in three perspectives: image-based affordance localization reasoning (Sec.~\ref{sec:image_reasoning}), zero-shot manipulation in simulation (Sec.~\ref{sec:manip_sim})) and in real-world settings (Sec.~\ref{sec:manip_real}) as shown in Fig.~\ref{fig:task_setting} and further conduct an ablation study on the effectiveness of mask-enhanced features, joint pose-centered learning, and compatibility with robot data for further validation (Sec.~\ref{sec:ablation}).

We compare our method against four baselines, namely VRB~\cite{bahl2023affordances}, RAM~\cite{kuangram}, MOKA~\cite{liu2024moka}, and RoboPoint~\cite{yuan2025robopoint}, which represent training-based, retrieve-and-transfer and VLM-based approaches respectively. As none of the baseline methods predict the manipulation pose directly, we utilize AnyGrasp~\cite{fang2023anygrasp} to produce grasp proposals and filter the final pose based on the contact point predicted by baselines as in~\cite{kuangram}. A rule-based mechanism is incorporated to avoid collisions with objects for our method accordingly. For manipulation tasks, prescribed post-contact waypoints are provided to facilitate task completion. The depth and object masks used for evaluation are obtained using the same pipeline as stated in Sec.~\ref{sec:data_pre}, if unavailable.

\begin{table}[t]
    \caption{Evaluation results on AGD20K dataset.}
    \vspace{-1em}
    \label{tab:imagedata}
    \begin{center}
    \begin{tabular}{p{2.5cm}|ccc}
    \toprule
    \bf Models & SR $\uparrow$ &  NSS $\uparrow$ &  DTM $\downarrow$ \\
    \midrule 
    RAM~\cite{kuangram} & 0.1824 & 0.1892 & 0.0689 \\
    RoboPoint~\cite{yuan2025robopoint} & 0.2138 & 0.2188 & 0.0508\\
    VRB~\cite{bahl2023affordances} & 0.2846 & 0.2557 & 0.2069 \\
    MOKA~\cite{liu2024moka} & 0.3711 & 0.3390 & \bf 0.0331 \\
    \midrule
    \bf \our{} (ours) & \bf 0.4403 & \bf 0.4083 &  0.0445\\
    \bottomrule
    \end{tabular}
    \vspace{-2em}
    \end{center}
\end{table}

\subsection{Image-Based Affordance Localization Reasoning}
\label{sec:image_reasoning}
To evaluate the precision of contact point localization, we conduct experiments on AGD20K~\cite{luo2022learning} datasets. Following~\cite{ju2024robo}, we select all the objects that are feasible for robotic manipulation and supplemented instances of drawer and cupboard categories following the same labeling procedure for comprehensive evaluation. Three metrics are reported for evaluation, including \textbf{Success Rate (SR)}, \textbf{Normalized Scanpath Saliency (NSS)}, and \textbf{Distance to Mask (DTM)}.

As shown in Tab.~\ref{tab:imagedata}, \our{} achieves a high success rate of 44.03\%, which is 18.6\% higher than the second-best method, MOKA, which leverages the strong reasoning capabilities of VLMs. This demonstrates the effectiveness of our method in 
localizing the appropriate contact point across categories, as shown in Fig.~\ref{fig:AGD20K}. Besides, \our{} also achieves higher NSS and similarly low DTM compared to MOKA, indicating that the predicted contact points are closer to the centers of the ground-truth masks. 

\subsection{Zero-shot Generalization across Tasks in Simulation}
\label{sec:manip_sim}

To evaluate the zero-shot generalization of \our{} across tasks, we evaluate the task success rates on RLBench~\cite{james2020rlbench}. 10 representative tasks from RLBench are selected as the evaluation set, including common object grasping, manipulation of task-relevant object regions, and articulated objects. Before evaluation, the manipulation regions of target objects are ensured to be fully visible by adjusting the camera viewpoint. Each task is evaluated with 25 episodes scored either 0 or 100, indicating failure or success in task execution. We report the average success rate for each task. The entire evaluation follows the testing pipeline of the modular approach as in~\cite{liu2024moka}, the task is first decomposed into a sequence of subtasks aligned with the robot’s meta-skills, based on the given task instruction and visual input. Each subtask is then executed by its corresponding module to generate action trajectories. We replace the contact-relevant meta-skill modules with affordance models to evaluate whether the 6-DoF pose predicted by the affordance model could support successful task execution.

As shown in Tab.~\ref{tab:rlbench}, our method outperforms all baselines on most of the tasks, yielding an average success rate of 64.8\%. Compared to our method, RoboPoint and VRB suffer from providing precise contact points for manipulation, leading to suboptimal performance on tasks, such as WaterPlants, which need to locate the contact point precisely on the handle of the watering can. RAM struggles to infer affordances for objects with diverse attributes (e.g., appearance), as its retrieval mechanism relies on object-level similarity within the affordance memory. Although MOKA achieves the second-best performance by leveraging the strong reasoning capabilities of VLMs, its performance is limited by AnyGrasp’s capability to generate grasp proposals in cluttered environments and under varying viewpoints. Moreover, the inherent randomness in VLMs' outputs can lead to failures in tasks that require highly consistent action patterns, such as StackBlocks. In contrast, our method achieves higher consistency between the predicted contact points and contact poses by jointly learning pose-centered affordances. Furthermore, by incorporating geometric features, it supports cross-category transfer and generalizes to objects with diverse shapes and appearances beyond the training categories.

\begin{table*}[htbp]
  \centering
  \caption{Success rate (\%) on RLBench Multi-Task setting.}
  \vspace{-1em}
    \begin{tabular}{l|cccccccccc|c}
    \toprule
    \bf Models &  \makecell{Pick\\Cup} &  \makecell{Stack\\Blocks} &  \makecell{Take\\Umbrella} &  \makecell{Water\\Plants}  & \makecell{Sweep\\Dirt} & \makecell{Close\\Laptop} &  \makecell{Open\\Drawer} &  \makecell{Close\\Drawer} &  \makecell{Open\\Microwave} &  \makecell{Close\\Microwave} & Avg.\\
    \midrule 
    RoboPoint~\cite{yuan2025robopoint} & 84 & 4 & 24 & 12 & 44 & 36 & 24 & 56 & 4 & 72 & 36.0\\
    VRB~\cite{bahl2023affordances} & 84 & 60 & 32 & 4 & 48 & 44 & 20 & 84 & 8 & 64 & 44.8\\
    RAM~\cite{kuangram} & 84 & 40 & 40 & 36 & 64 & 44 & 16 & 60 & 8 & 60 & 45.2\\
    MOKA~\cite{liu2024moka} & 84 & 36 & 32 & 20 & 48 & \textbf{52} & 28 & \textbf{88} & 8 & 72 & 46.8\\
    \midrule
    \bf \our{} (ours)~ & \textbf{88} & \textbf{72} & \textbf{88} & \textbf{44} & \textbf{68} & \textbf{52} & \textbf{36} & 84 & \textbf{32} & \textbf{84} & \textbf{64.8}\\
    \bottomrule
    \end{tabular}
  \label{tab:rlbench}
\end{table*}

\begin{figure}[t]
    \centering
    \includegraphics[width=0.5\textwidth]{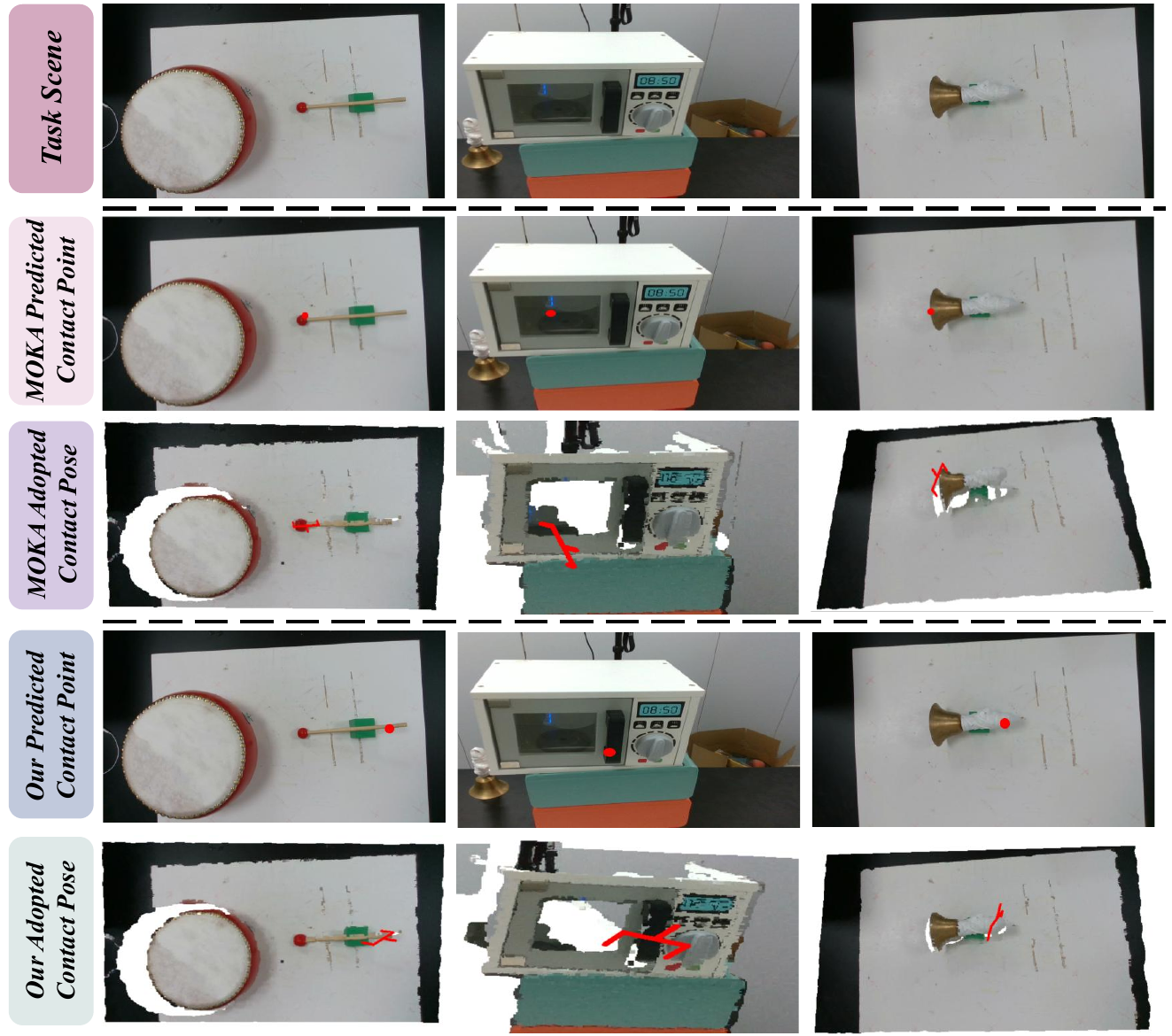}
    \vspace{-2em}
    \caption{Qualitative comparison of our model’s predicted contact points and poses with MOKA in real-world settings.}
    \vspace{-1em}
    \label{fig:qualitative_manip}
\end{figure}

\subsection{Real-World Experiments}
\label{sec:manip_real}
In the real-world setting, we carefully design 9 tasks involving interactions with diverse household objects, encompassing articulated objects (e.g., drawers), objects with function-specific regions (e.g., drumsticks), and deformable objects (e.g., cloth). We employ a Franka Emika robotic arm with a parallel gripper and utilize an on-hand RealSense D435i camera to capture the RGB-D image of the scene. We conduct 10 trials for each task, under different spatial arrangements and with varying object instances, to validate the generalization capability across tasks and categories.

As shown in Tab.~\ref{tab:real}, \our{} achieves an 83.3\% success rate on average among all 9 tasks, 24.9\% higher than the second-best method, RAM, which yields conclusions consistent with the experimental results on the simulation platform. Although the strong real-world grasp generation performance of AnyGrasp enables baseline methods to accomplish some pick-and-place tasks even with inaccurate contact point predictions, they struggle to support tasks that require precise prediction of contact point within task-specific regions (e.g., PlayDrum). Some qualitative results are provided in Fig.~\ref{fig:qualitative_manip} to further illustrate the results.

\begin{table*}[htbp]
  \centering
  \caption{Success rate (\%) on 9 tasks in Real-world setting.}
  \vspace{-1em}
    \begin{tabular}{l|ccccccccc|c}
    \toprule
    \bf Models &  \makecell{Put\\Into} &  \makecell{Tidy\\Table} &  \makecell{Pour\\Water} & \makecell{Shake\\Bell} & \makecell{Play\\Drum}  & \makecell{Open\\Drawer} &  \makecell{Close\\Drawer} & \makecell{Open\\Microwave} & \makecell{Close\\Microwave} & Avg.\\
    \midrule 
    MOKA~\cite{liu2024moka} & 90 & \bf 100 & \bf 80 & 50 & 60 & 50 & 70 & 10 & 80  & 65.6\\
    RAM~\cite{kuangram} & \bf 100 & 80 & 60 & 50 & 30 & \bf 70 & 70 & \bf 60 & 80 &  66.7\\
    \midrule
    \bf \our{} (ours)~ & \textbf{100} & 90 & 70 & \textbf{80} & \textbf{100} & \bf 70 & \textbf{80} & \textbf{60} & \textbf{100} & \textbf{83.3}\\
    \bottomrule
    \end{tabular}
  \label{tab:real}
\end{table*}

\subsection{Ablation Study}
\label{sec:ablation}

As in Tab.~\ref{tab:ablation}, we conduct detailed ablation experiments on a subset of manipulation tasks on RLBench, which consist of articulated object and function-specific region manipulation. These tasks are selected as they are more representative of evaluating how effectively the predicted pose-centered affordances support tasks that require precise manipulation of function-specific object regions. Each task is evaluated with 25 episodes under the same settings as in Sec.~\ref{sec:manip_sim}.

\begin{table}[htbp]
  \centering
  \caption{Ablation results on 5 representative tasks evaluated on success rate (\%). AT01: SweepDirt, AT02: OpenDrawer, AT03: CloseDrawer, AT04: OpenMicrowave, AT05: CloseMicrowave.}
  \vspace{-1em}
    \begin{tabular}{l|c@{\hspace{5pt}}c@{\hspace{5pt}}c@{\hspace{5pt}}c@{\hspace{5pt}}c|c}
    \toprule
    \bf Models &  \makecell{AT01} &  \makecell{AT02} &  \makecell{AT03} & \makecell{AT04} & \makecell{AT05} & Avg.\\
    \midrule 
    \bf \our{} (ours) & \bf 68 & \bf 36 & 84 & 32 & 84 & 60.8\\
    \midrule
    w/ robot data & 60 & \bf 36 & 92 & \bf 40 & \bf 96 & \bf 64.8 \\
    w/ AnyGrasp & 60 & 16 & 92 & 8 & 72 & 49.6 \\
    w/o masked features & 12 & 0 & \bf 96 & 28 & 80 & 43.2 \\
    \bottomrule
    \end{tabular}
  \vspace{-1em}
  \label{tab:ablation}
\end{table}

\textbf{Effectiveness of mask-enhanced features.}
We compare our model trained with and without mask-enhanced features. For the model without mask-enhanced features, the RGB-D encoder encodes tokens from the original RGB-D frame directly to represent the scene, which allows us to evaluate the effectiveness of mask-enhanced features on the model’s performance in capturing task-relevant object regions. As shown in the last row of Tab.~\ref{tab:ablation}, the success rate decreases markedly, especially for tasks requiring precise contact point predictions, such as OpenDrawer, which underscores the importance of incorporating mask-enhanced features for accurate pose-centered affordance prediction.

\textbf{Effectiveness of joint pose-centered learning.}
To further evaluate the effectiveness of the pose-centered affordance learning paradigm, which jointly predicts contact points and contact poses, we compare our model with and without AnyGrasp. For the model with AnyGrasp, we replace the output of 6-DoF pose with the grasp pose obtained by filtering the grasp proposals generated by AnyGrasp based on the predicted contact point from our model. As shown in the third row of Tab.~\ref{tab:ablation}, the performance of our model outperforms the model with AnyGrasp, which demonstrates the advantage of jointly learning contact points and poses for producing reliable and consistent manipulation actions.

\textbf{Compatibility with robot data.}
Beyond training only on human demonstrations, we also investigate whether \our{} can leverage robot demonstrations to further improve the performance of pose-centered affordance prediction. We collect 2K data samples based on DROID~\cite{khazatsky2024droid} for evaluation, each comprising an RGB-D frame, the interacting object mask, the contact point, and the corresponding contact pose. As shown in the second row of Tab.~\ref{tab:ablation}, incorporating robot data improves performance among most of the tasks, demonstrating that \our{} is compatible with robot demonstrations and can effectively benefit from additional robotic experience.

\section{CONCLUSION AND FUTURE WORK}
In this work, we propose \textbf{\our{}}, a pose-centered affordance prediction model that jointly predicts the contact points and poses given task instructions. To reduce collection costs, we devise \textbf{\ourdata{}}, which automatically recovers scene-level 3D information and extracts pose-centered affordances from unlabeled human demonstrations. Extensive experiments show that \our{} achieves higher accuracy in contact point prediction, stronger consistency between contact points and manipulation poses, and generalization across tasks and categories. Ablation studies further validate the effectiveness of mask-enhanced features, the pose-centered affordance learning paradigm, and compatibility with robot data. Future work will extend our approach to cross-embodiment and larger datasets, enabling more versatile, robust manipulation across diverse objects and scenarios. 









\bibliographystyle{./IEEEtran}
\bibliography{./IEEEabrv,./IEEEexample}

@book{gibson2014ecological,
  title={The ecological approach to visual perception: classic edition},
  author={Gibson, James J},
  year={2014},
  publisher={Psychology press}
}

@article{gibson1966senses,
  title={The senses considered as perceptual systems.},
  author={\vspace{0mm}Gibson, James Jerome},
  year={1966},
  publisher={Houghton Mifflin}
}

@inproceedings{qian2024affordancellm,
  title={Affordancellm: Grounding affordance from vision language models},
  author={Qian, Shengyi and Chen, Weifeng and Bai, Min and Zhou, Xiong and Tu, Zhuowen and Li, Li Erran},
  booktitle={Proceedings of the IEEE/CVF Conference on Computer Vision and Pattern Recognition},
  year={2024}
}

@inproceedings{tang2024uad,
  title={UAD: Unsupervised Affordance Distillation for Generalization in Robotic Manipulation},
  author={Tang, Yihe and Huang, Wenlong and Wang, Yingke and Li, Chengshu and Yuan, Roy and Zhang, Ruohan and Wu, Jiajun and Fei-Fei, Li},
  booktitle={CoRL 2024 Workshop on Mastering Robot Manipulation in a World of Abundant Data}
}

@article{ma2024glover,
  title={Glover: Generalizable open-vocabulary affordance reasoning for task-oriented grasping},
  author={Ma, Teli and Wang, Zifan and Zhou, Jiaming and Wang, Mengmeng and Liang, Junwei},
  journal={arXiv preprint arXiv:2411.12286},
  year={2024}
}

@inproceedings{bahl2023affordances,
  title={Affordances from human videos as a versatile representation for robotics},
  author={Bahl, Shikhar and Mendonca, Russell and Chen, Lili and Jain, Unnat and Pathak, Deepak},
  booktitle={Proceedings of the IEEE/CVF Conference on Computer Vision and Pattern Recognition},
  year={2023}
}

@inproceedings{kuangram,
  title={RAM: Retrieval-Based Affordance Transfer for Generalizable Zero-Shot Robotic Manipulation},
  author={Kuang, Yuxuan and Ye, Junjie and Geng, Haoran and Mao, Jiageng and Deng, Congyue and Guibas, Leonidas and Wang, He and Wang, Yue},
  booktitle={Conference on Robot Learning},
  year={2025},
}

@inproceedings{ju2024robo,
  title={Robo-abc: Affordance generalization beyond categories via semantic correspondence for robot manipulation},
  author={Ju, Yuanchen and Hu, Kaizhe and Zhang, Guowei and Zhang, Gu and Jiang, Mingrun and Xu, Huazhe},
  booktitle={European Conference on Computer Vision},
  year={2024},
}

@article{ma2025glover++,
  title={GLOVER++: Unleashing the Potential of Affordance Learning from Human Behaviors for Robotic Manipulation},
  author={Ma, Teli and Zheng, Jia and Wang, Zifan and Gao, Ziyao and Zhou, Jiaming and Liang, Junwei},
  journal={arXiv preprint arXiv:2505.11865},
  year={2025}
}

@inproceedings{liu2024moka,
  title={Moka: Open-vocabulary robotic manipulation through mark-based visual prompting},
  author={Liu, Fangchen and Fang, Kuan and Abbeel, Pieter and Levine, Sergey},
  booktitle={First Workshop on Vision-Language Models for Navigation and Manipulation at ICRA 2024},
  year={2024}
}

@inproceedings{yuan2025robopoint,
  title={RoboPoint: A Vision-Language Model for Spatial Affordance Prediction in Robotics},
  author={Yuan, Wentao and Duan, Jiafei and Blukis, Valts and Pumacay, Wilbert and Krishna, Ranjay and Murali, Adithyavairavan and Mousavian, Arsalan and Fox, Dieter},
  booktitle={Conference on Robot Learning},
  year={2025},
  organization={PMLR}
}

@inproceedings{chen2025vidbot,
  title={VidBot: Learning Generalizable 3D Actions from In-the-Wild 2D Human Videos for Zero-Shot Robotic Manipulation},
  author={Chen, Hanzhi and Sun, Boyang and Zhang, Anran and Pollefeys, Marc and Leutenegger, Stefan},
  booktitle={Proceedings of the Computer Vision and Pattern Recognition Conference},
  year={2025}
}

@article{xu2025a0,
  title={A0: An affordance-aware hierarchical model for general robotic manipulation},
  author={Xu, Rongtao and Zhang, Jian and Guo, Minghao and Wen, Youpeng and Yang, Haoting and Lin, Min and Huang, Jianzheng and Li, Zhe and Zhang, Kaidong and Wang, Liqiong and others},
  journal={arXiv preprint arXiv:2504.12636},
  year={2025}
}

@article{khazatsky2024droid,
  title={Droid: A large-scale in-the-wild robot manipulation dataset},
  author={Khazatsky, Alexander and Pertsch, Karl and Nair, Suraj and Balakrishna, Ashwin and Dasari, Sudeep and Karamcheti, Siddharth and Nasiriany, Soroush and Srirama, Mohan Kumar and Chen, Lawrence Yunliang and Ellis, Kirsty and others},
  journal={arXiv preprint arXiv:2403.12945},
  year={2024}
}

@inproceedings{liu2022hoi4d,
  title={Hoi4d: A 4d egocentric dataset for category-level human-object interaction},
  author={Liu, Yunze and Liu, Yun and Jiang, Che and Lyu, Kangbo and Wan, Weikang and Shen, Hao and Liang, Boqiang and Fu, Zhoujie and Wang, He and Yi, Li},
  booktitle={Proceedings of the IEEE/CVF Conference on Computer Vision and Pattern Recognition},
  year={2022}
}

@article{fang2023anygrasp,
  title={Anygrasp: Robust and efficient grasp perception in spatial and temporal domains},
  author={Fang, Hao-Shu and Wang, Chenxi and Fang, Hongjie and Gou, Minghao and Liu, Jirong and Yan, Hengxu and Liu, Wenhai and Xie, Yichen and Lu, Cewu},
  journal={IEEE Transactions on Robotics},
  year={2023},
  publisher={IEEE}
}

@article{tang2025foundationgrasp,
  title={Foundationgrasp: Generalizable task-oriented grasping with foundation models},
  author={Tang, Chao and Huang, Dehao and Dong, Wenlong and Xu, Ruinian and Zhang, Hong},
  journal={IEEE Transactions on Automation Science and Engineering},
  year={2025},
  publisher={IEEE}
}

@inproceedings{luo2022learning,
  title={Learning affordance grounding from exocentric images},
  author={Luo, Hongchen and Zhai, Wei and Zhang, Jing and Cao, Yang and Tao, Dacheng},
  booktitle={Proceedings of the IEEE/CVF conference on computer vision and pattern recognition},
  year={2022}
}

@article{james2020rlbench,
  title={Rlbench: The robot learning benchmark \& learning environment},
  author={James, Stephen and Ma, Zicong and Arrojo, David Rovick and Davison, Andrew J},
  journal={IEEE Robotics and Automation Letters},
  volume={5},
  number={2},
  pages={3019--3026},
  year={2020},
  publisher={IEEE}
}

@inproceedings{delitzas2024scenefun3d,
  title={Scenefun3d: Fine-grained functionality and affordance understanding in 3d scenes},
  author={Delitzas, Alexandros and Takmaz, Ayca and Tombari, Federico and Sumner, Robert and Pollefeys, Marc and Engelmann, Francis},
  booktitle={Proceedings of the IEEE/CVF Conference on Computer Vision and Pattern Recognition},
  year={2024}
}

@inproceedings{do2018affordancenet,
  title={Affordancenet: An end-to-end deep learning approach for object affordance detection},
  author={Do, Thanh-Toan and Nguyen, Anh and Reid, Ian},
  booktitle={2018 IEEE international conference on robotics and automation (ICRA)},
  year={2018}
}

@inproceedings{geng2023partmanip,
  title={Partmanip: Learning cross-category generalizable part manipulation policy from point cloud observations},
  author={Geng, Haoran and Li, Ziming and Geng, Yiran and Chen, Jiayi and Dong, Hao and Wang, He},
  booktitle={Proceedings of the IEEE/CVF Conference on Computer Vision and Pattern Recognition},
  year={2023}
}

@inproceedings{mo2021where2act,
  title={Where2act: From pixels to actions for articulated 3d objects},
  author={Mo, Kaichun and Guibas, Leonidas J and Mukadam, Mustafa and Gupta, Abhinav and Tulsiani, Shubham},
  booktitle={Proceedings of the IEEE/CVF International Conference on Computer Vision},
  year={2021}
}

@inproceedings{damen2018scaling,
  title={Scaling egocentric vision: The epic-kitchens dataset},
  author={Damen, Dima and Doughty, Hazel and Farinella, Giovanni Maria and Fidler, Sanja and Furnari, Antonino and Kazakos, Evangelos and Moltisanti, Davide and Munro, Jonathan and Perrett, Toby and Price, Will and others},
  booktitle={Proceedings of the European conference on computer vision (ECCV)},
  year={2018}
}

@article{hoque2025egodex,
  title={EgoDex: Learning Dexterous Manipulation from Large-Scale Egocentric Video},
  author={Hoque, Ryan and Huang, Peide and Yoon, David J and Sivapurapu, Mouli and Zhang, Jian},
  journal={arXiv preprint arXiv:2505.11709},
  year={2025}
}

@article{nair2022r3m,
  title={R3m: A universal visual representation for robot manipulation},
  author={Nair, Suraj and Rajeswaran, Aravind and Kumar, Vikash and Finn, Chelsea and Gupta, Abhinav},
  journal={arXiv preprint arXiv:2203.12601},
  year={2022}
}

@article{xu2024flow,
  title={Flow as the cross-domain manipulation interface},
  author={Xu, Mengda and Xu, Zhenjia and Xu, Yinghao and Chi, Cheng and Wetzstein, Gordon and Veloso, Manuela and Song, Shuran},
  journal={arXiv preprint arXiv:2407.15208},
  year={2024}
}

@inproceedings{bharadhwaj2024towards,
  title={Towards generalizable zero-shot manipulation via translating human interaction plans},
  author={Bharadhwaj, Homanga and Gupta, Abhinav and Kumar, Vikash and Tulsiani, Shubham},
  booktitle={2024 IEEE International Conference on Robotics and Automation (ICRA)},
  year={2024},
  organization={IEEE}
}

@article{srirama2024hrp,
  title={Hrp: Human affordances for robotic pre-training},
  author={Srirama, Mohan Kumar and Dasari, Sudeep and Bahl, Shikhar and Gupta, Abhinav},
  journal={arXiv preprint arXiv:2407.18911},
  year={2024}
}

@article{chang2023look,
  title={Look ma, no hands! agent-environment factorization of egocentric videos},
  author={Chang, Matthew and Prakash, Aditya and Gupta, Saurabh},
  journal={Advances in Neural Information Processing Systems},
  year={2023}
}

@article{chen2021learning,
  title={Learning generalizable robotic reward functions from" in-the-wild" human videos},
  author={Chen, Annie S and Nair, Suraj and Finn, Chelsea},
  journal={arXiv preprint arXiv:2103.16817},
  year={2021}
}

@inproceedings{shaw2023videodex,
  title={Videodex: Learning dexterity from internet videos},
  author={Shaw, Kenneth and Bahl, Shikhar and Pathak, Deepak},
  booktitle={Conference on Robot Learning},
  year={2023},
  organization={PMLR}
}

@article{wang2023mimicplay,
  title={Mimicplay: Long-horizon imitation learning by watching human play},
  author={Wang, Chen and Fan, Linxi and Sun, Jiankai and Zhang, Ruohan and Fei-Fei, Li and Xu, Danfei and Zhu, Yuke and Anandkumar, Anima},
  journal={arXiv preprint arXiv:2302.12422},
  year={2023}
}

@article{haldar2025point,
  title={Point policy: Unifying observations and actions with key points for robot manipulation},
  author={Haldar, Siddhant and Pinto, Lerrel},
  journal={arXiv preprint arXiv:2502.20391},
  year={2025}
}

@article{bharadhwaj2024gen2act,
  title={Gen2act: Human video generation in novel scenarios enables generalizable robot manipulation},
  author={Bharadhwaj, Homanga and Dwibedi, Debidatta and Gupta, Abhinav and Tulsiani, Shubham and Doersch, Carl and Xiao, Ted and Shah, Dhruv and Xia, Fei and Sadigh, Dorsa and Kirmani, Sean},
  journal={arXiv preprint arXiv:2409.16283},
  year={2024}
}

@article{chi2023diffusion,
  title={Diffusion policy: Visuomotor policy learning via action diffusion},
  author={Chi, Cheng and Xu, Zhenjia and Feng, Siyuan and Cousineau, Eric and Du, Yilun and Burchfiel, Benjamin and Tedrake, Russ and Song, Shuran},
  journal={The International Journal of Robotics Research},
  pages={02783649241273668},
  year={2023},
  publisher={SAGE Publications Sage UK: London, England}
}

@article{ze20243d,
  title={3d diffusion policy: Generalizable visuomotor policy learning via simple 3d representations},
  author={Ze, Yanjie and Zhang, Gu and Zhang, Kangning and Hu, Chenyuan and Wang, Muhan and Xu, Huazhe},
  journal={arXiv preprint arXiv:2403.03954},
  year={2024}
}

@article{ke20243d,
  title={3d diffuser actor: Policy diffusion with 3d scene representations},
  author={Ke, Tsung-Wei and Gkanatsios, Nikolaos and Fragkiadaki, Katerina},
  journal={arXiv preprint arXiv:2402.10885},
  year={2024}
}

@inproceedings{liang2024skilldiffuser,
  title={Skilldiffuser: Interpretable hierarchical planning via skill abstractions in diffusion-based task execution},
  author={Liang, Zhixuan and Mu, Yao and Ma, Hengbo and Tomizuka, Masayoshi and Ding, Mingyu and Luo, Ping},
  booktitle={Proceedings of the IEEE/CVF Conference on Computer Vision and Pattern Recognition},
  year={2024}
}

@article{janner2022planning,
  title={Planning with diffusion for flexible behavior synthesis},
  author={Janner, Michael and Du, Yilun and Tenenbaum, Joshua B and Levine, Sergey},
  journal={arXiv preprint arXiv:2205.09991},
  year={2022}
}

@article{hu2024metric3d,
  title={Metric3d v2: A versatile monocular geometric foundation model for zero-shot metric depth and surface normal estimation},
  author={Hu, Mu and Yin, Wei and Zhang, Chi and Cai, Zhipeng and Long, Xiaoxiao and Chen, Hao and Wang, Kaixuan and Yu, Gang and Shen, Chunhua and Shen, Shaojie},
  journal={IEEE Transactions on Pattern Analysis and Machine Intelligence},
  year={2024},
  publisher={IEEE}
}

@inproceedings{liu2024grounding,
  title={Grounding dino: Marrying dino with grounded pre-training for open-set object detection},
  author={Liu, Shilong and Zeng, Zhaoyang and Ren, Tianhe and Li, Feng and Zhang, Hao and Yang, Jie and Jiang, Qing and Li, Chunyuan and Yang, Jianwei and Su, Hang and others},
  booktitle={European conference on computer vision},
  year={2024},
}

@article{ravi2024sam,
  title={Sam 2: Segment anything in images and videos},
  author={Ravi, Nikhila and Gabeur, Valentin and Hu, Yuan-Ting and Hu, Ronghang and Ryali, Chaitanya and Ma, Tengyu and Khedr, Haitham and R{\"a}dle, Roman and Rolland, Chloe and Gustafson, Laura and others},
  journal={arXiv preprint arXiv:2408.00714},
  year={2024}
}

@article{team2024gemini,
  title={Gemini 1.5: Unlocking multimodal understanding across millions of tokens of context},
  author={Team, Gemini and Georgiev, Petko and Lei, Ving Ian and Burnell, Ryan and Bai, Libin and Gulati, Anmol and Tanzer, Garrett and Vincent, Damien and Pan, Zhufeng and Wang, Shibo and others},
  journal={arXiv preprint arXiv:2403.05530},
  year={2024}
}

@inproceedings{shan2020understanding,
  title={Understanding human hands in contact at internet scale},
  author={Shan, Dandan and Geng, Jiaqi and Shu, Michelle and Fouhey, David F},
  booktitle={Proceedings of the IEEE/CVF conference on computer vision and pattern recognition},
  year={2020}
}

@inproceedings{pavlakos2024reconstructing,
  title={Reconstructing hands in 3d with transformers},
  author={Pavlakos, Georgios and Shan, Dandan and Radosavovic, Ilija and Kanazawa, Angjoo and Fouhey, David and Malik, Jitendra},
  booktitle={Proceedings of the IEEE/CVF Conference on Computer Vision and Pattern Recognition},
  year={2024}
}

@inproceedings{xiao2024spatialtracker,
  title={Spatialtracker: Tracking any 2d pixels in 3d space},
  author={Xiao, Yuxi and Wang, Qianqian and Zhang, Shangzhan and Xue, Nan and Peng, Sida and Shen, Yujun and Zhou, Xiaowei},
  booktitle={Proceedings of the IEEE/CVF Conference on Computer Vision and Pattern Recognition},
  year={2024}
}

@inproceedings{zhou2019continuity,
  title={On the continuity of rotation representations in neural networks},
  author={Zhou, Yi and Barnes, Connelly and Lu, Jingwan and Yang, Jimei and Li, Hao},
  booktitle={Proceedings of the IEEE/CVF conference on computer vision and pattern recognition},
  year={2019}
}

@inproceedings{yin2025dformerv2,
  title={Dformerv2: Geometry self-attention for rgbd semantic segmentation},
  author={Yin, Bo-Wen and Cao, Jiao-Long and Cheng, Ming-Ming and Hou, Qibin},
  booktitle={Proceedings of the Computer Vision and Pattern Recognition Conference},
  year={2025}
}

@article{kitaev2020reformer,
  title={Reformer: The efficient transformer},
  author={Kitaev, Nikita and Kaiser, {\L}ukasz and Levskaya, Anselm},
  journal={arXiv preprint arXiv:2001.04451},
  year={2020}
}

\end{document}